\documentclass{article}

\usepackage[preprint]{neurips_2023}

\usepackage[utf8]{inputenc} 
\usepackage[T1]{fontenc}    
\usepackage{hyperref}       
\usepackage{url}            
\usepackage{booktabs}       
\usepackage{amsfonts}       
\usepackage{nicefrac}       
\usepackage{microtype}      
\usepackage[dvipsnames]{xcolor}         

\definecolor{myblue}{RGB}{16, 72, 245}
\hypersetup{
    colorlinks=true,
    linkcolor=Maroon,
    citecolor=blue,      
    urlcolor=Blue, 
    }
\usepackage{graphicx, wrapfig} 
\usepackage{subcaption} 
\usepackage{amsmath} 
\usepackage{algorithm}
\usepackage{algpseudocode}
\usepackage{slashbox}
\usepackage{multirow}
\usepackage{pgfplots}
\pgfplotsset{compat=1.18}
\usepackage{graphbox}
\usepackage{todonotes}  
\usepackage{tabularx}
\usepackage[export]{adjustbox}

\usepackage{xr}
\makeatletter

\newcommand*{\addFileDependency}[1]{
\typeout{(#1)}
\@addtofilelist{#1}
\IfFileExists{#1}{}{\typeout{No file #1.}}
}\makeatother

\newcommand{\mean}[2]{\mathbb{E}_{#2} \! \left[ #1 \right]}
\newcommand{\vect}[1]{\boldsymbol{#1}}
\newcommand\norm[2]{{\left\lVert#1\right\rVert}_{#2}}

\pgfplotsset{
   standard/.style={
    axis x line=middle,
    axis y line=middle,
    axis z line=middle,
    enlarge x limits=0.15,
    enlarge y limits=0.15,
    enlarge x limits=0.15,
    every axis x label/.style={at={(current axis.right of origin)},anchor=north west},
    every axis y label/.style={at={(current axis.above origin)},anchor=north east},
    every axis z label/.style={at={(current axis.above origin)},anchor=south}
}
}

\title{The statistical thermodynamics of generative diffusion models: Phase transitions, symmetry breaking and critical instability}

\author{%
 Luca ~Ambrogioni$^{1,2}$ \\
$^1$Radboud University \\
$^2$Donders Institute for Brain, Cognition and Behaviour\\ 
\texttt{luca.ambrogioni@donders.ru.nl}\\ 
}

\begin{document}

\maketitle

\begin{abstract}
    Generative diffusion models have achieved spectacular performance in many areas of machine learning and generative modeling. While the fundamental ideas behind these models come from non-equilibrium physics, variational inference and stochastic calculus, in this paper we show that many aspects of these models can be understood using the tools of equilibrium statistical mechanics. Using this reformulation, we show that generative diffusion models undergo second-order phase transitions corresponding to symmetry breaking phenomena. We show that these phase-transitions are always in a mean-field universality class, as they are the result of a self-consistency condition in the generative dynamics. We argue that the critical instability that arises from the phase transitions lies at the heart of their generative capabilities, which are characterized by a set of mean-field critical exponents. Finally, we show that the dynamic equation of the generative process can be interpreted as a stochastic adiabatic transformation that minimizes the free energy while keeping the system in thermal equilibrium. 
\end{abstract}

\section{Introduction}
Generative modeling is a sub-field of machine learning concerned with the automatic generation of structured data such as images, videos and written language \citep{bond2021deep}. Generative diffusion models \citep{sohl2015deep}, also known as score-based models, form a class of deep generative models that have demonstrated high performance in image  \citep{ho2020denoising, song2021scorebased}, sound \citep{chen2020wavegrad, kong2020diffwave, liu2023audioldm} and video generation \citep{ho2022video, singer2022make}. Diffusion models were first introduced in analogy with the physics of non-equilibrium statistical physics. The fundamental idea is to formalize generation as the probabilistic inverse of a \emph{forward stochastic process} that gradually turns the target distribution $\phi(\vect{y})$ into a simple base distribution such as Gaussian white noise \citep{sohl2015deep, song2020denoising}. Recently, several works suggested that many of the dynamical properties of generative diffusion models can be understood using concepts such as spontaneous symmetry breaking \citep{raya2023spontaneous, biroli2023generative, biroli2024dynamical}, and phase transitions \citep{biroli2024dynamical, sclocchi2024phase}. These theoretical and experimental results suggest a deep connection between generative diffusion and equilibrium phenomena. 

In this paper, we outline a conceptual reformulation of generative diffusion models in the language of equilibrium statistical physics. We begin by defining a family of Boltzmann distributions over the noise-free states, which are interpreted as (unobservable) microstates during the diffusion process. In this picture, the Boltzmann weights are given by the conditional distributions of the noiseless data given the noisy state. We obtain a self-consistent equation of state for the system, which corresponds to the fixed-point equation of the generative dynamics. Moreover, we show that generative diffusion models can undergo second-order phase transitions of the mean-field type, corresponding the the generative spontaneous symmetry breaking phenomena fist discussed in \citep{raya2023spontaneous} and further studied in \cite{biroli2024dynamical, li2024critical} and in \cite{sclocchi2024phase}. Finally, we show that this mean-field theory can be seen as the thermodynamic limit of a multi-site system of coupled replicas. Based on this results, we derive a variant of the generative diffusion equations as the Brownian dynamics of a 'particle' coupled on a large densely connected systems of replicated microstates, which offers a possible generalization of diffusion models beyond mean-field theory. 

\section{Contributions and Related Work}
The main novel theoretical contributions of this paper are in the characterization of mean-field critical phase transitions in generative diffusion models, and its extension beyond mean-field theory. While the paper contains novel results, its aim is also pedagogical, as we wish to provide a self-consistent introduction for physicists to the study of generative diffusion. As such, we report known formulas and results from the literature, including the analysis scheme used in \citep{lucibello2024exponential} and \cite{biroli2024dynamical} for the analysis of memorization phenomena and the equivalence results for modern Hopfield networks given in \citep{ambrogioni2023search}.  Several of these formulas can also be found in recent work on stochastic localization \citep{el2022sampling, huang2024sampling}, which has been shown to offer an elegant generalization of generative diffusion processes \citep{montanari2023sampling, benton2024nearly, alaoui2023sampling}. In particular, the Boltzmann distributions given in Eq.~\ref{eq: boltzmann equation} is equivalent to the tilted distributions given in \citep{el2022sampling}.  

\section{Preliminaries on generative diffusion models}
The goal of diffusion modeling is to sample from a potentially very complex target distribution $\phi(\vect{y})$, which we model as the initial initial boundary condition of a (forward) stochastic process that removes structure by injecting white noise. In order to simplify the derivations, here we assume the forward process to be a mathematical Brownian motion. Other forward processes are more commonly used in the applied literature, such as the variance preserving process (e.g. a non-stationary Olsten-Uhlenbeck process) \citep{song2021scorebased}. However, most of the qualitative thermodynamic properties are shared between these models. The mathematical Brownian motion is defined by the following Langevin equation:
\begin{equation} \label{eq: forward}
    \vect{x}(t + \text{d} t) =  \vect{x}(t) +  ~\sigma ~ \vect{w}(t) \sqrt{\text{d} t}
\end{equation}
where $\text{d} t$ is an infinitesimal increment, $\sigma$ is the instantaneous standard deviation of the stochastic input and $w(t)$ is a standard Gaussian white noise process. The marginal probabilities defined by Eq.~\ref{eq: forward} with $\phi(\vect{y})$ as initial boundary condition can be expressed analytically as follows:
\begin{equation} \label{eq: marginals}
    p_t(\vect{x}) = \frac{1}{\sqrt{2 \pi t \sigma^2}} \mean{e^{-\frac{\norm{\vect{x} - \vect{y}}{2}^2}{2 t \sigma^2}}}{\vect{y} \sim \phi}
\end{equation}
where the expectation is taken with respect to the target distribution $\phi(\vect{y})$. A generative model can then be obtained by "inverting" Eq.~\ref{eq: forward}. The inverse equation is  
\begin{equation} \label{eq: reversed dynamycs}
    \vect{x}(t - \text{d} t) =  \vect{x}(t) - \sigma^2 \nabla \log{p_t(\vect{x})} +  ~\sigma ~ w(t) \sqrt{\text{d} t}~,
\end{equation}
which can be shown to give the same marginal distributions in Eq.~\ref{eq: marginals} if the process is initialized with appropriately scaled white-noise \citep{anderson1982reverse}. The function $\nabla \log{p_t(\vect{x})}$ is known as the score in the literature. If the score is available for all values of $\vect{x}$ and $t$, we can then sample from $\phi(\vect{y})$ by integrating Eq.~\ref{eq: reversed dynamycs} using numerical methods.

\subsection{Training diffusion models as denoising autoencoders}
\begin{figure}
\centering 
\includegraphics[width=13cm]{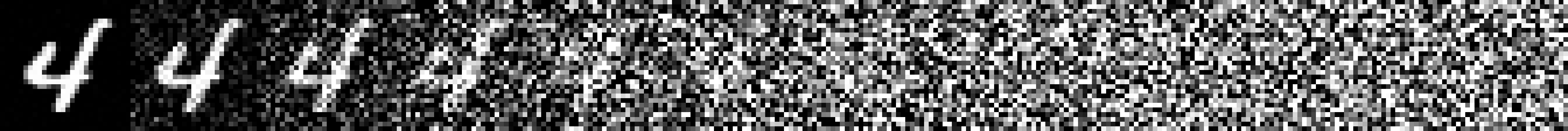}
\caption{Generative process for a digit taken from the MNIST dataset.} \label{fig: mnist process}
\end{figure}
While the score of the target distribution is generally not available analytically, a deep network can be trained to approximate it using a from a large set of samples \citep{song2020denoising}. We refer to the network as a vector valued function $\vect{f}(\vect{x}(t), t)$. Deep networks are parameterized by a large number of weights and biases. However, since we are not interested in the details of the specific parameterization, here we will report the functional loss:
\begin{equation}\label{eq: autoencoder loss}
    \mathcal{L}\left[f(\cdot, \cdot) \right] = \int_0^{t_{\text{end}}} \mean{\frac{1}{2}\norm{\frac{\vect{x}(t) - \vect{y}}{\sqrt{t \sigma^2}} - \vect{f}(\vect{x}(t), t)}{2}^2}{\vect{y}, \vect{x}(t)} \text{d} \Psi(t)~,
\end{equation}
where $\Psi(t)$ is a cumulative distribution with support in $(0, t_{\text{end}})$  and $\vect{x}(t)$ is sampled conditional on $\vect{y}$ using the propagator of the forward Langevin equation. Note that ${\vect{x}(t) - \vect{y}}/{\sqrt{t \sigma^2}}$ is simply the total noise added up to time $t$, which implies that the network learns how to predict the noise that corrupted the input data. The score function $\nabla \log{p_t(\vect{x})}$ can then be obtained from the optimized network $\mathbf{f}^* = \text{argmin}~ \mathcal{L}\left[f \right]$ using the following formula \citep{anderson1982reverse}:
\begin{equation} \label{eq: score from net}
    \nabla \log{p_t(\vect{x})} \approx -(t \sigma^2)^{-{1/2}} f^*(\vect{x}(t), t)~.
\end{equation}
In other words, the score is proportional to the optimal estimate of the noise given the noise-corrupted state. Therefore, once the network is trained to minimize Eq.~\ref{eq: autoencoder loss}, synthetic samples can be generated by sampling $\vect{x}(t_\text{end})$ from the boundary noise, computing the score using Eq.~\ref{eq: score from net} and integrating backward Eq.~\ref{eq: reversed dynamycs} using numerical methods. An example of this generative dynamics for a network trained on natural images is shown in Fig.~\ref{fig: mnist process}. 

\section{Diffusion models as systems in equilibrium}
The starting point for any model in statistical mechanics is the definition of the relevant microstates. In statistical physics, the microstate of a system is usually assumed to be an unobservable quantity. Given that we can observe a noise-corrupted data $\vect{x}^*(t)$, in a diffusion model the most obvious unobservable quantity of interest is the noise-free initial state $\vect{x}^*(0) = \vect{y}^*$. The next step is to define a Hamiltonian function on the set of microstates. We can do this by considering the conditional probability of the data $\vect{y}$ given a noisy state $\vect{x}_t$:
\begin{equation} \label{eq: boltzmann equation}
        p(\vect{y} \mid \vect{x}, t) = \frac{1}{p_t(\vect{x}) \sqrt{2 \pi t \sigma^2}} e^{-\frac{\norm{\vect{x} - \vect{y}}{2}^2}{2 t \sigma^2} + \log{\phi(\vect{y})}} = \frac{1}{Z(\vect{x},t)} e^{-\mathcal{H}(\vect{y}; \vect{x}, t)}~,
\end{equation}
which we can interpret as a Boltzmann distribution over the microstates $\vect{y}$ with Hamiltonian 
\begin{equation} \label{eq: hamiltonian}
    \mathcal{H}(\vect{y}; \vect{x}, t) = (t \sigma^2)^{-1} \left(\frac{1}{2}\norm{\vect{y}}{2}^2 - \vect{x} \cdot \vect{y}\right) - \log{\phi(\vect{y})}
\end{equation}
and partition function 
\begin{equation}
    Z(\vect{x}, t) = \int e^{-\mathcal{H}(\vect{y}; \vect{x}, t)} \text{d} \vect{y}~.
\end{equation}
The statistical properties of this ensemble determine the score function, which can be expressed as a Boltzmann average:
\begin{equation}
    \nabla \log{p_t(\vect{x})} = \beta(t) (\langle \vect{y} \rangle_{t,\vect{x}} - \vect{x})~,
\end{equation}
where $\beta(t) = (t \sigma^2)^{-1}$ and
\begin{equation}
    \langle \vect{y} \rangle_{t,\vect{x}} = \int \vect{y} ~\frac{e^{-\mathcal{H}(\vect{y}; \vect{x}, t)}}{Z(\vect{x}, t)} \text{d} \vect{y}~.
\end{equation}
Intuitively, this equation tells us that the score vector directs the system towards the posterior average $\langle \vect{y} \rangle_{t,\vect{x}}$. As we shall see, studying the thermodynamics determined by these weights will allow us to understand several important qualitative and quantitative features of the generative dynamics. For example, as we will see in later sections, after a 'condensation' phase transition the score will only depend on a small number of data-points, which can be detected by studying the concentration of the weights on a sub-exponential number of microstates. 

The thermodynamic system defined by Eq.~\ref{eq: boltzmann equation} does not have a true temperature parameter. However, the quantity $\sigma t$ plays a very similar role to temperature in classical statistical mechanics. Moreover, in the Hamiltonian given by Eq~.\ref{eq: hamiltonian}, the dynamic variable $\vect{x}$ is analogous to the external field term in magnetic systems, which can bias the distribution of microstates towards the patterns 'aligned' in its direction. We can imagine $\vect{x}$ as being a "slower" thermodynamic variable that interacts (adiabatically) with the statistics of the microstates. 

\subsection{Example 1: Two deltas}
Most of the complexity of the generative dynamics comes from the target distribution $\phi(\vect{y})$. However, simple toy models can be used to draw general insights that often generalize to complex target distributions. A simple but informative example is given by the following target
\begin{equation}
    \phi(y) = \frac{1}{2} \left(\delta(y + 1) + \delta(y - 1)\right)
\end{equation}
where $y$ is equal to either $-1$ or $1$ with probability $1/2$. Assuming the binary constraint, this results in the following diffusion Hamilton
\begin{equation} \label{eq: deltas hamiltonian}
    \mathcal{H}(\vect{y}; \vect{x}, t) = (t \sigma^2)^{-1} \left(\frac{1}{2}y^2 - x y \right) - \log2
\end{equation}
and the partition function 
\begin{equation}
    Z(x, t) = e^{\beta/2} \cosh{\left(\beta(t) x \right)}~.
\end{equation}
where $\beta(t) = \sigma^2 t$. Note that this is the same partition function of the Curie-Weiss model of ferromagnetism, which suggests a connection with mean-field theory. 

\subsection{Example 2: Discrete dataset}
In real application, generative diffusion models are trained on a large but finite dataset $D = \{\vect{y}_1,\dots,\vect{y}_N\}$. Sampling from this dataset correspond to the target distribution
\begin{equation} \label{eq: patterns hamiltonian}
    \phi(\vect{y}) = \frac{1}{N} \sum_{j=1}^N \delta(\vect{y} - \vect{y}_j)~.
\end{equation}
If the data-points are all normalized so as to have norm equal to one, this results in the partition function
\begin{equation}
    Z(\vect{x}, t, N) = \frac{e^{-\beta(t)/2}}{N} \sum_{j=1}^N e^{\beta(t) ~\vect{x} \cdot \vect{y}_j}~.
\end{equation}
This partition function will play a central role in the random-energy analysis of the model, which can be used to study the finite sample thermodynamics.

\subsection{Example 3: Hyper-spherical manifold}
Since datasets are always finite, in practice every trained generative diffusion model  corresponds to the discrete model outlined in the previous subsection. However, fitting the dataset exactly leads to a model that can only reproduce the memorized training data. Instead, the hope is that the trained network will generalize and interpolate the samples, thereby approximately recovering the true distribution of the sampled data. Very often, this distribution will span a lower-dimensional manifold embedded in the ambient space. 

A simple toy model of data defined in a manifold is the hyper-spherical model introduced in \citep{raya2023spontaneous}: 
\begin{equation}
    Z(\vect{x}, t, d) = \frac{e^{-\beta(t)/2}}{V(d-1)} \int_{S(d-1)} e^{-\beta(t) \vect{x} \cdot \vect{y}} \text{d} \vect{y}~.
\end{equation}
where $S(d-1)$ denotes a $d-1$-dimensional hyper-sphere centered at zero with volume $V(d-1)$. The "two delta" model is a special case of this model for an ambient dimension equal to one. As we will see in further section, this data distribution is very tractable in the infinite dimensional (i.e. thermodynamic) limit as it converges to a distribution of normalized Gaussian variables, which removes the quadratic terms in the Hamiltonian. 

\subsection{Example 4: Diffused Ising model}
While most of the formulas presented in this manuscript have a very close analogy with formulas in statistical physics, there are some subtle interpretative differences that could create confusion in the reader. To clarify these issues, we will discuss the \emph{diffused Ising model}, which will provide a bridge between the two views. Consider a diffusion model with a target distribution supported on $d$-dimensional vectors $\vect{y}$ with entries in the set $\{-1, 1\}$. The log-probability of the target distribution is defined by the following formula:
\begin{equation}
    \log{\phi(\vect{y})} = - \frac{1}{2 T} \sum_{j \neq k} y_j y_k W_{jk} + c~,
\end{equation}
where W is a symmetric coupling matrix, $T$ is a temperature parameter and $c$ is a constant. Up to constants, this is of course the log-probability of an Ising model without the external field term. From Eq.~\ref{eq: hamiltonian}, up to constant terms, we obtain the following Hamiltonian for the diffusion model:
\begin{equation} \label{eq: diffused Ising hamiltonian}
    \mathcal{H}(\vect{y}; \vect{x}, t, T) = - \beta(t) \vect{x} \cdot \vect{y} + \frac{1}{2 T} \sum_{j \neq k} y_j y_k W_{jk}~,
\end{equation}
which is almost identical to the Hamiltonian of an Ising model coupled to a location-dependent external field $\vect{x}$. Nevertheless, the quantity $\beta(t) = (t \sigma^2)^{-1}$, which we loosely interpreted as a "inverse temperature", does not divide the coupling part of the Hamiltonian, which results in a radically different behavior. In fact, $t \sigma^2$ only modulates the susceptibility to the field term and it therefore does not radically alter the phase of the model, which depends on the Ising temperature parameter $T$. Instead, the interesting phase transition of the diffusion models is a consequence of the self-consistency relation in Eq.~\ref{eq: self-consistency}, which characterizes the branching of the fixed-points of the generative stochastic dynamics. From the point of view of statistical physics, Eq.~\ref{eq: self-consistency} can be seen as the result of a mean-field approximation, where the average magnetization is coupled to the external field. However, it is important to keep in mind that, in a diffusion model, this mean-field approach does not represent the coupling between individual sites, which, as Eq.~\ref{eq: diffused Ising hamiltonian} shows, are instead statistically coupled by the interaction terms in the Hamiltonian. Instead, it can be seen as an idealized mean-field interaction between infinitely many copies of the whole system. In general, the value of $T$ will change the properties of the diffusion model, as the system transitions from its low temperature to its high temperature phase. The dependency of the diffusion dynamics on this transition have been studied in \cite{biroli2023generative}.

\section{Free energy, magnetization and order parameters}
\begin{figure}
\centering 
\includegraphics[width=14cm]{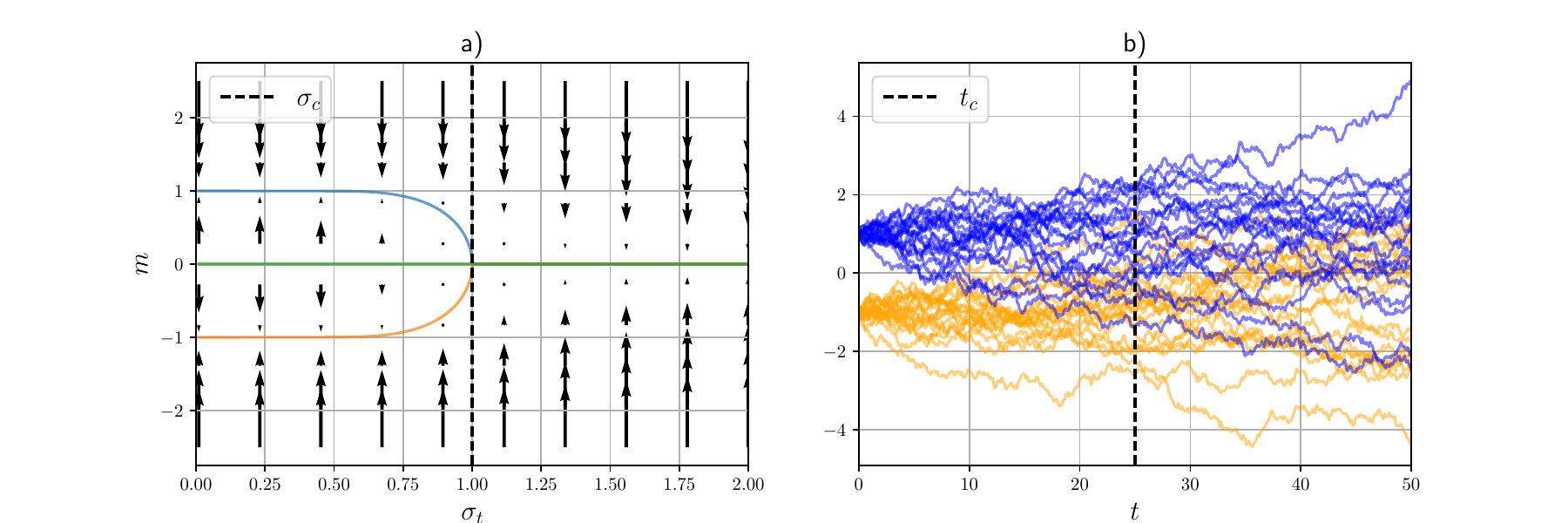}
\caption{Visualization of a phase transition in a simple diffusion model (two deltas). a) Order parameter paths and (regularized) free energy gradients. The dashed line denotes the critical value of $\sigma_t = \sqrt{t} \sigma_0$. b) Forward process. The dashed line denotes time critical time. } \label{fig: phase transition}
\end{figure}
Using our interpretation of $\beta(t) = (t \sigma^2)^{-1}$ as the inverse temperature parameter, we can define the Helmholtz free energy as follows
\begin{equation}
    \mathcal{F}(\vect{x}, t) = - \beta^{-1}(t) \log{Z(\vect{x}, t)}~.
\end{equation}
The expected value of the pattern $\vect{y}$ given $\vect{x}$ can then be expressed the gradient of the free energy with respect to $\vect{x}$:
\begin{equation} \label{eq: expectation}
    \langle \vect{y} \rangle_{t,\vect{x}} = - \nabla \mathcal{F}(\vect{x}, t)~.
\end{equation}
This formula suggests an analogy between diffusion models and magnetic systems in statistical physics. The noisy state $\vect{x}$ can be interpreted as an external magnetic field, which induces the state of "magnetization" $\langle \vect{y} \rangle_{t,\vect{x}}$. In this analogy, a diffusion model is magnetized when its distribution is biased towards a sub-set of the possible microstates. 

In physics, the 'external field' variable $\vect{x}$ is usually assumed to be controlled by the experimenter. On the other hand, in generative diffusion models $\vect{x}$ is a dynamic variable that, under the reversed dynamics, is itself attracted towards $\langle \vect{y} \rangle_{t,\vect{x}}$ by the drift term:
\begin{equation}
    \nabla \log{p_t(\vect{x})} = \beta(t) \left(\langle \vect{y} \rangle_{t,\vect{x}} - \vect{x}\right)
\end{equation}
In other words, if we ignore the effect of the dispersion term, the state of the system is driven towards self-consistent points where $\vect{x}$ is equal to $\langle \vect{y} \rangle_{t,\vect{x}}$. It is therefore interesting to study the self-consistency equation
\begin{equation} \label{eq: self-consistency}
    \vect{m}(\vect{h}, t) = -\nabla \mathcal{F}(\vect{m}(\vect{h}, t) + \vect{h}, t)~,
\end{equation}
which defines the self-consistent solutions $\vect{m}(t)$ where the state is identical to the expected value. In the equation, we introduced a perturbation term $\vect{h}$, which will allow us to study how the systems react to perturbations. For $\vect{h} = 0$, the equation can be equivalently re-expressed as the fixed point equation of the reversed drift:
\begin{equation}
    \nabla \log{p_t(\vect{m}(\vect{0},t))} = \vect{0}~.
\end{equation}
For $t \rightarrow \infty$, this equation admits the single "trivial" solution $\vect{m} = \langle \vect{y} \rangle_0$, where $\langle \cdot\rangle_0$ denotes expectation with respect to the target distribution $\phi(\vect{y})$. In analogy with magnetic systems, we can interpret $\vect{m}(\vect{h}, t)$ as an order parameter and this equation as a thermodynamic equation of state. This analogy suggests that $\vect{m}(\vect{h}, t)$ can be interpreted as a 'spontaneous magnetization' of the system. From this point of view, we can conceptualize the generative process as a form of self-consistent spontaneous symmetry breaking, where the system aligns with one of the many possible target points. In the following sections, we will formalize this insight by characterizing the critical behavior of this system. 

Readers familiar with statistical physics will recognize that Eq.~\ref{eq: self-consistency} is formally identical to the self-consistency conditions used in the mean-field approximation, where the external field term in one location is assumed to be determined by the magnetization of all other locations. However, it is important to note that in the case of a diffusion model, this self-consistent coupling is not approximate, as it is a natural consequence of the dynamics. Nevertheless, the formal analogy implies that the thermodynamics of generative diffusion models is formally identical to the thermodynamics of mean-field models. 

\subsection{The susceptibility matrix}
In the physics of magnetic systems, the magnetic susceptibility matrix determines how much the different magnetization components are sensitive to the components of the external magnetic field. Similarly, in diffusion models we can define a susceptibility matrix:
\begin{equation}
\label{eq: susceptibility}
    \chi_{ij}(\vect{x}, t) = \frac{\partial m_i}{\partial h_j} \bigg\rvert_{\vect{h} = 0}~,
\end{equation}
which tells us how sensitive the expected value is to changes in the noisy-state $\vect{x}(t)$. The susceptibility matrix is helpful in interpreting the dynamics of the generative denoising process as it informs us on how random fluctuations in each component of the state $\vect{x}(t)$ are propagated to the other components. For example, in the context of image generation, a random fluctuation of "green" at the bottom of an image can propagate to the rest, originating the image of a forest. 

The susceptibility matrix can be re-expressed in temrms of connected correlation matrix (i.e. the covariance matrix) of the microstates under the Boltzmann distribution
\begin{equation}
    \chi_{ij}(\vect{x}, t) = - \beta(t) \left( \langle \vect{y} \vect{y}^T \rangle_{\vect{x}} - \langle \vect{y} \rangle_{t,\vect{x}} \langle \vect{y}^T \rangle_{\vect{x}} \right) = - \beta(t) C_{ij}(\vect{x},t)
\end{equation}

We can now express the Jacobi matrix of the score function as follows
\begin{equation}
    J_{ij}(\vect{x}) = \beta(t) \left( \chi_{ij}(\vect{x}, t)  - \delta_{j,k}\right) = - \beta(t) \delta_{j,k} -  \beta(t)^2 C_{ij}(\vect{x},t)~.
\end{equation}
\begin{figure}
\centering 
\includegraphics[width=13cm]{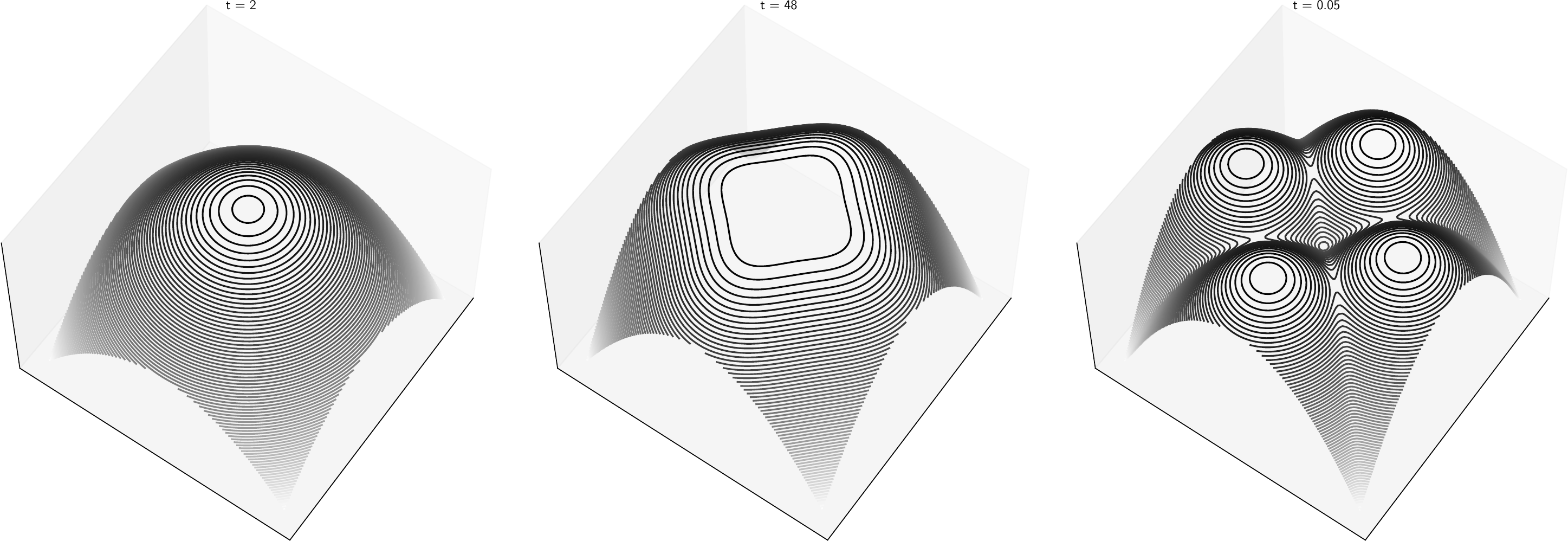}
\caption{Negative free energy of a "four delta" 2d diffusion model for different values of the time variable. The target points are at $(1,0)$, $(0,1)$, $(-1,0)$ and $(0,-1)$.} \label{fig: free energy}
\end{figure}

\section{Phase transitions and symmetry breaking}
A spontaneous symmetry breaking happens when the trivial solution for the order parameter branches into multiple solutions at some critical time of $t_c$. This corresponds to the onset of multi-modality in the regularized free energy $\tilde{\mathcal{F}}(\vect{x}, t)$, as visualized in Fig.~\ref{fig: free energy} for a "four deltas" model. In thermodynamic systems, the symmetry breaking corresponds to a (second order) phase transition, which in this case can be detected by the divergence of several state variables around the critical point. 

The presence of one (or multiple) phase transitions in diffusion models depends on the target distribution $\phi(\vect{y})$. The simplest example is given by the "two deltas", which corresponds to the process visualized in Fig.~\ref{fig: phase transition} a. Using this distribution, we obtain the self-consistency equation
\begin{equation} \label{eq: z1 equation}
    m = \tanh{\left(\frac{m}{t \sigma^2} \right)}
\end{equation}
The solutions of this equation are shown in Fig.~\ref{fig: phase transition} b, together with the gradient of the regularized free energy, where we can see the branching of the solutions and the singular behavior around a critical point. Eq.~\ref{eq: z1 equation} is identical to the mean-field self-consistency equation of an Ising model, from which we can deduce that the critical scaling of this simple generative diffusion models shares its universality class. For example, by Taylor expansion of Eq.\ref{eq: z1 equation} around $t_c = 1$, we see that
\begin{equation}
    m \sim (-\tau)^{1/2} 
\end{equation}
with $\tau = t - t_c$, which is valid for $t$ smaller than $t_c$. 

\subsection{Generation and critical instability}
As shown in \citep{raya2023spontaneous}, spontaneous symmetry breaking phenomena play a central role in the generative dynamics of diffusion models. Consider the simple "two deltas" model. For $t > >t_c$, the dynamics is mean-reverting towards a unique fixed-point $\vect{m}(\vect{0}, t)$. Around $t = t_c$, the order parameter splits into thee "branches", an unstable one corresponding to the mean of the target distribution and two stable ones corresponding to the two target points. Importantly, at the critical point the susceptibility defined in Eq.~\ref{eq: susceptibility} diverges, implying that the system becomes extremely reactive to fluctuations in the noise. This instability is determined by the critical exponents $\delta$ and $\gamma$ and , defined by the relations
\begin{equation}
    h_j \propto m_j^{\delta_j}~,
\end{equation}
and
\begin{equation}
    \chi_{ij} \propto (-\tau)^{\gamma_{ij}}~.
\end{equation}
Note that, in the general case the critical exponents can be different for different coordinates and matrix entries. These divergences give rise to something we refer to as \emph{critical generative instability}. We conjecture that the diversity of the generated samples crucially depends on a proper sampling of this critical region.

\section{Generation as as an adiabatic free energy descent process}
 So far, we characterized the thermodynamic state of diffusion model at time $t$ by its Boltzmann distribution. The dynamics of the system can now be recovered as a form of (stochastic) free energy minimization:
\begin{equation} \label{eq: free energy dynamycs}
    \vect{x}(t - \text{d}t) = \vect{x}(t) - \beta(t) \nabla \tilde{\mathcal{F}}(\vect{x}, t) \text{d} t + \sigma \vect{w}(t) \sqrt{\text{d} t}~,
\end{equation}
where $\tilde{\mathcal{F}}$ is the free-energy plus a free potential term:
\begin{equation}
    \tilde{\mathcal{F}}(\vect{x}, t) = \mathcal{F}(\vect{x}, t) + V(\vect{x})
\end{equation}
where $V(\vect{x}) = \frac{1}{2} \norm{\vect{x}}{2}^2$. This can be seen as a form of adiabatic approximation, where the dynamics of the 'slow' variable $\vect{x}$ is obtained by assuming that the system is maintained in thermal equilibrium along the diffusion trajectory. The symmetry breaking can now be detected as a change of shape in the regularized free energy, which transitions from a convex shape with a single global minimum to a more complex shape with potentially several meta-stable points (see Fig.~\ref{fig: free energy}). The reformulation of the dynamics in term of the gradient of a free energy allows us to interpret generative diffusion models as a kind of energy-based machine learning models \citep{lecun2006tutorial}, as discussed in \citep{hoover2023memory} and \citep{ambrogioni2023search}. The main difference is that the (free) energy is not learned directly but it is instead implicit in the learned score function. The connection suggests potential connection with the free energy principle in theoretical neuroscience, which is used to characterize the stochastic dynamics of biological neural systems \citep{friston2010free}. 

\section{Beyond mean-field theory: A multi-site 'generative bath' model}
There results given in the previous sections suggest that generative diffusion models can be seen as a mean-field limit of a model with replicated microstates on $K$ 'sites' coupled through long-range interactions. We denote the microstate in the $\mu$-th site as $\vect{y}^{\mu}$. Consider the following multi-site Hamiltonian:
\begin{equation} \label{eq: hamiltonian}
    \mathcal{H}_K[\vect{y}^{1:K}; \vect{h}] =  \beta \left( \frac{w}{2} \sum_{\mu, \nu, ~\mu \neq \nu}^K \vect{y}^{\mu} \cdot \vect{y}^{\nu} - \sum_j^K \vect{y}^{\mu} \cdot \vect{h} \right) - \sum_j \log{\psi(\vect{y}^{\mu})}
\end{equation}
where $\vect{y}^{1:K}$ denotes a multi-site configuration and $w$ is coupling weight parameter and $\vect{h}$ is an external field term. The re-normalized marginal energy $- \log{\psi(\vect{y}^{\mu})}$ is defined to fulfil the following constraint:
\begin{equation}
    \lim_{\beta \rightarrow \infty} \frac{1}{Z_K(\beta, \vect{h})} e^{-\beta \mathcal{H}_K[\vect{y}^{1:K}; \vect{h}]} = \phi(\vect{y}^1) \prod_{\mu=2}^K \delta(\vect{y}^{\mu} - \vect{y}^1)~.
\end{equation}
In general, we have that $\log{\psi(\vect{y})} = \log{\phi(\vect{y})}$ when the possible microstates have a constant euclidean norm, while it will involve additional re-normalization terms in the general case. This condition ensures that the multi site coupling only affects the correlation between different sites (leading to perfect alignment for $\beta \rightarrow \infty$) while preserving the limiting marginal distributions. We refer to this thermodynamic system as a \emph{generative bath}.

In the model, different replications of the microstates $\vect{y}_j$ (i.e. the noise-free data) exert mutual attractive couplings. Generation can be seen as a spontaneous symmetry breaking that the system undergoes when the temperature (i.e. the time) decreases, since at low temperatures all the microstates in all sites will align on the same pattern, resulting in a coherent observable average
\begin{equation} \label{eq: coupled replicated hamiltonian}
    \bar{\vect{y}} = \frac{1}{K} \sum_\mu^K \vect{y}^{\mu}~.
\end{equation}
In the thermodynamic limit ($K \rightarrow 0$), the model converges to the self-consistent mean-field model discussed in the previous sections. This allows us to conceptualize the self-consistency condition implicit in the fixed-point equation of generative diffusion model as the result of an ideal multi-site coupling, which could result in new forms of neural implementations. This conceptualization opens the door for possible non-mean-field generalizations of generative diffusion characterized by short-range interactions, or disordered generalizations with random interactions. However, it is not clear if these extensions will have practical value.

Depending on the choice of the target distribution $\phi(\vect{y})$, the coupled model in Eq.~\ref{eq: coupled replicated hamiltonian} reduces to well known models in statistical mechanics such as the fully connected Ising model for the two-deltas distribution and the classical Heisenberg model for spherical distributions. However, the model becomes substantially more complex under more realistic distributions of the data. 

\subsection{Brownian dynamics in a 'generative bath'}
The Hamiltonian defined in Eq.~\ref{eq: coupled replicated hamiltonian} specifies an equilibrium system that, in the thermodynamic limit, shares the same self-consistent criticality of the fixed-point equation of generative diffusion models. In this section, we will derive from first principles a generative stochastic dynamics similar to the generative equation given in Eq.~\ref{eq: reversed dynamycs}. The idea is to consider a 'Brownian particle' $\vect{x}(t)$ coupled to the multi-site system of microstates. We define the random force as
\begin{equation} \label{eq: random force}
    F(\vect{x}, t) = \frac{1}{H} \sum_{\mu}^H \left(\vect{y}^{\mu}  - \vect{x}\right)~,
\end{equation}
where $H$ is the number of sites that are coupled to $\vect{x}$. In contrast to the distribution in Eq.~\ref{eq: boltzmann equation}, we assume that $\vect{x}$ does not exert any effect on the equilibrium systems itself, which instead undergoes symmetry breaking events due to its own internal coupling between sites. In other word, in this formulation the state $\vect{x}(t)$ is now passively controlled by the statistical fluctuations in the equilibrium system. 

If we assume that the force in Eq.~\ref{eq: random force} is applied at each infinitesimal time interval and that its time-scale is much faster that the motion of $\vect{x}(t)$, the Brownian dynamics follows the following (reversed) Langevin equation
\begin{equation} \label{eq: free energy dynamycs}
    \vect{x}(t - \text{d}t) - \vect{x}(t) = - \left(\langle \bar{\vect{y}} \rangle_{t,\vect{x} \rightarrow 0} - \vect{x}(t)\right) \frac{\text{d} t}{t} + \frac{1}{\sqrt{H}} B(\vect{x}, t) \vect{w}(t) \sqrt{\frac{\text{d} t}{t}}~,
\end{equation}
where $B(\vect{x}, t) = C^{1/2}(\vect{x})$ is a matrix square root of the pure state covariance matrix:
\begin{equation}
C(\vect{x}) = \left(\langle \bar{\vect{y}} \bar{\vect{y}}^T \rangle_{t, \vect{x} \rightarrow 0} - \langle \bar{\vect{y}} \rangle_{t, \vect{x} \rightarrow 0}\langle \bar{\vect{y}}^T \rangle_{t, \vect{x} \rightarrow 0} \right)~.
\end{equation}
The $1/t$ scaling in the differential are introduced to have the reversed diffusion ends at the finite time $t = 0$, which is equivalent to a logarithmic change of coordinate in the time variable. 

The Boltzmann expectation is taken with respect to the multi-site ensemble given in Eq.~\ref{eq: coupled replicated hamiltonian} in the limit of a vanishing external field alligned to $\vect{x}$. This is done in order to isolate the appropriate 'pure state' from the Boltzmann average, since after a spontaneous symmetry breaking only one branch of the distribution should affect the particle. In fact, after a symmetry breaking phase transition, the Boltzmann distribution splits into two or more modes corresponding to the possible states with broken symmetry. 

\subsection{The two delta model revisited}
In the "two deltas" model, the Hamiltonian of the generative bath" is just
\begin{equation} \label{eq: hamiltonian}
    \mathcal{H}_K[y^{1:K}; h] =  \beta \left( \frac{w}{2} \sum_{\mu, \nu, ~\mu \neq \nu}^K y^{\mu} y^{\nu} - h \sum_j^K y^{\mu} \cdot \right) 
\end{equation}
with the restriction that that $y^{\mu} \in \{\pm 1 \}$. This is simply the Hamiltonian of a fully connected Ising model with uniform coupling weights. In the thermodynamic limit $K \rightarrow \infty$, the model therefore reduces to the mean-field Curie-Weiss model that we have already discussed. In this case, the pure state magnetisation are the stable solutions $m_t$ of the self-consistency equation $m = \tanh{(m+h)/(t \sigma^2)}$, which is identically equal to zero for $t \sigma^2 > 1$ and it has two branches that cannot be expressed in closed-form in the low-temperature regime. The instantaneous variance of the Brownian generative dynamics is given by $T \partial m_t/ \partial h$, which is equal to $1/(1 - (t \sigma^2)^{-1})$ in the high temperature phase and to $T (1 - m_t^2)/m_t^2$ in the low temperature phase. Note that the variance diverges at $t \sigma^2 = 1$ due to the critical phase transition and that it vanishes for $t \rightarrow 0$ as the system fully aligns in one of the two possible pure states. 

\section{Associative memory and Hopfield networks}
We will now move back to the standard mean-field formulation of generative diffusion and discuss its connection with associative memory networks. Associative memory networks are energy-based learning systems that can store patterns (i.e. memories) as meta-stable states of a parameterized energy function \citep{hopfield1982neural, abu1985information, krotov2023new}. There is a substantial body of literature on the thermodynamic properties of associative memory networks \citep{strandburg1992phase, volk1998phase, marullo2020boltzmann}. The original associative memory networks, also known as \emph{Hopfield networks}, are defined by the energy function $E(\vect{x}) = \frac{1}{2} \vect{x}^T W \vect{x}$ under the constraints of binary entries for the state vector. In a Hopfield network, a finite number of training patterns $\vect{y}_j$ are encoded into a weight matrix $W = \sum_{j} \vect{y}_j \vect{y}_j^T$, which usually gives the correct minima when the number of patterns is on the order of the dimensionality. Associative memory networks can reach much higher capacity by using exponential energy function \citep{krotov2016dense, demircigil2017model, krotov2023new}. For example, \citep{ramsauer2020hopfield} introduces the use of the following function
\begin{equation} \label{eq: modern Hopfield}
    E(\vect{x}) = -\beta^{-1} \log{\left(\sum_j e^{\beta \vect{x} \cdot \vect{y}_j} \right)} + \frac{1}{2} \norm{\vect{x}}{2}^2~,
\end{equation}
which can be proven to provide exponential scaling of the capacity and it is related to the transformers architectures used in large language models \citep{ramsauer2020hopfield}. By inspection of Eq.~\ref{eq: modern Hopfield}, we can see that this energy function is equivalent to the regularized Helmholtz  free energy of a diffusion models trained on a mixture of delta distributions \citep{ambrogioni2023search}:
\begin{equation}
    \phi(\vect{y}) = \sum_j \delta(\vect{y} - \vect{y}_j)~,
\end{equation}
which gives a free energy with the same fixed-point structure of Eq.~\ref{eq: modern Hopfield} at the zero temperature limit. Note that, while the dynamics of a diffusion model does not necessarily act as an optimizer in the general case, the free energy is exactly optimized when $\phi(\vect{y})$ is a sum of delta functions, making the dynamics of the model exactly equivalent to the optimization of Eq.~\ref{eq: modern Hopfield} for $\beta \rightarrow \infty$. Given this connection, most of the results presented in this paper can be re-stated for associative memory networks. However, generative diffusion models are more general as they can target arbitrary  mixtures of continuous and singular distributions. As we shall see in the next section, the modern Hopfield Hamiltonian plays a crucial role in studying finite sample effects such as data memorization \citep{lucibello2024exponential}. 

\section{The random energy thermodynamic of diffusion models on sampled datasets}
As we saw in the previous sections, using the target density $\phi(\vect{y})$, we can define an Hamiltonian function, and consequently equilibrium thermodynamic system, that captures the statistical properties of the corresponding generative diffusion dynamics. From the point of view of machine learning, this corresponds to a denoising network perfectly trained on an infinitely large dataset. Consequently, this analysis misses some important properties that arise when the models are trained on finite datasets sampled independently from $\phi(\vect{y})$. In particular, the exact model cannot capture the phenomenon of memorization (overfitting), where the model fails to generalize beyond its training set. 

We can analyze this regime using the statistical physics of disordered systems, where the quenched partition function depends on $N$ randomly sampled training points:
\begin{equation}
    Z_N(\vect{x}, \beta(t)) = \frac{1}{N} \sum_{j=1}^N e^{-\beta(t) \left( \frac{1}{2}\norm{\vect{y}_j}{2}^2 + \vect{x} \cdot \vect{y}_j \right)}~,
\end{equation}
where $\vect{y}_j \sim \phi(\vect{y})$. The idea is that we can understand the finite sample properties of diffusion models by studying how the partition function and other thermodynamic quantities fluctuate as a function of the random sampling. This is analogous to the physics of glasses, where the thermodynamic properties depend randomly on the (disordered) structure of a piece of material. By defining $E_j(\vect{x}) = \frac{1}{2}\norm{\vect{y}_j}{2}^2 -\vect{x} \cdot \vect{y}_j$, we can re-write this quenched partition function as a random energy model (REM):
\begin{equation}\label{eq: random partition}
    Z_N(\vect{x}, \beta(t)) = \frac{1}{N} \sum_{j=1}^N e^{-\beta(t) E_j(\vect{x})}~.
\end{equation}
where the distribution of the energy levels $E_j(\vect{x})$ depends on the "external field" $\vect{x}$. The thermodynamic of this model is well known for simple distributions over the energy, and can be studied in more complex cases using the replica method \citep{mezard1987spin}. 

\subsection{Memorization as 'condensation'}
The random-energy analysis of a diffusion model is very useful to study the memorization phenomenon, where the diffusion trajectories collapse on the individual sampled datapoints instead of spreading to the underlying target distribution. In machine learning jargon, this phenomenon is known as overfitting. It is clear that a model perfectly trained on a finite dataset (i.e. a model that perfectly reproduces the empirical score) fully memorizes the data without any generalization for $t \rightarrow 0$. However, these over-fitted models can still exhibit generalization for finite values of $t$ since the noise level can be too high to distinguish the individual training points. It was recently discovered that, in the thermodynamic limit, generalization and memorization are demarcated by a disordered symmetry breaking \citep{lucibello2024exponential, biroli2024dynamical}. This is a so-called condensation phenomenon, where the probability measures transition from being spread out over an exponential number of configurations to being concentrated on a random small (sub-exponential) set. During condensation, the score is determined by a relatively small number of non-vanishing Boltzmann weight, which eventually direct the dynamics towards one of the training points.

From our theoretical framework, we can find this 'critical time' $t_{\text{cond}}$ by studying the random partition function given in Eq.~\ref{eq: random partition} in the thermodynamic limit. This can be done by evaluating the 'participation ration'
\begin{equation}
    Y_N(\vect{x}, t) = \frac{Z_N(\vect{x}, 2\beta(t))}{\left(Z_N(\vect{x}, \beta(t)) \right)^2}~,
\end{equation}
where $1/Y_N(\vect{x}, t)$ gives a rough count of the  number of configurations with non-vanishing probability. In the REM theory, the participation ration can be used to detect a condensation phase transition, as in this case its value becomes identically equal to zero for $\beta > \beta_c$ in a non-analytic manner. 

Consider the simple case where we sampled $N = 2^M$ the data-points from the distribution $\phi(\vect{y})$ , which is uniform over a $d$-dimensional hyper-sphere with radius $r = \nu \sqrt{M/2d}$, where $\nu$ is a parameter that regulates the standard deviation of the data. In this case, up to an irrelevant constant shift, the random energy becomes
\begin{equation}
    E_j(\vect{x}) = - \vect{x} \cdot \vect{y}_j~.
\end{equation}
For high values of $d$, the distribution of $\vect{x} \cdot \vect{y}_j$ is approximately a centered Gaussian with variance $\norm{x}{2}^2 \nu^2 M/2$ and, which $\phi(\vect{y})$ is spherically symmetric, it does not depend on the direction of $\vect{x}$. We can therefore approximately re-write the expression as a standard REM model:
\begin{equation}
    Z_N(\vect{x}, \beta(t)) = \frac{1}{2^M} \sum_{j=1}^{2^M} e^{-\tilde{\beta} \tilde{E}_j}~.
\end{equation}
where $\tilde{E}_j \sim \mathcal{N}(0, M/2)$ and $\tilde{\beta} = \nu \beta(t) \norm{\vect{x}}{2}$. Note that this formulation is not exact at the thermodynamic limit as it ignores the residual effect of non-Gaussianity in the distribution. A more rigourous analysis can be found in \citep{lucibello2024exponential} (Appendix C) in the context of modern Hopfield networks. 

For the standard REM model, we know that the critical condensation temperature is at $\beta_c = 2 \sqrt{\log{2}}$ (see \cite{mezard1987spin}), which leads to the critical time  
\begin{equation}
    t_{\text{cond}}(\vect{x}) = \frac{\nu \norm{x}{2}}{2 \sigma^2 \sqrt{\log{2}}}~.
\end{equation}
This formula was first derived in \cite{lucibello2024exponential} for associative memory models. It can be shown that, at the limit $M \rightarrow \infty$, the expected participation ratio is
\begin{equation}
    \mean{Y(\vect{x}, t)}{} = 
    \left\{
    \begin{array}{lr}
        0, & \text{if } \beta(t) \leq \beta_c\\
        1 - \frac{\beta_c}{\beta(t)}, & \text{if } \beta(t) > \beta_c
    \end{array}
\right\}~.
\end{equation}
This allows us to estimate the number of data-points with non-negligible weight in the score as $n(\vect{x}) \approx 1/Y(\vect{x}, t)$. Note that $t_c$ does not predict the collapse on an individual data-point, as that would require $Y(\vect{x}, t) \approx 1$, which happens at an earlier time. However, it identifies a critical transition in the score function $\nabla \log{p(\vect{x})}$ that demarcates the beginning of a memorization phase. In this phase, the system undergoes further symmetry breaking phase transitions, which break the symmetry between the data-points with non-vanishing weights. The critical behavior of these transitions is analogous to the "two deltas" model.

\begin{figure}[ht!]
\centering
\begin{minipage}[b]{.353\textwidth}
  \begin{minipage}[b]{\textwidth}
    \centering
    \includegraphics[width=\textwidth]{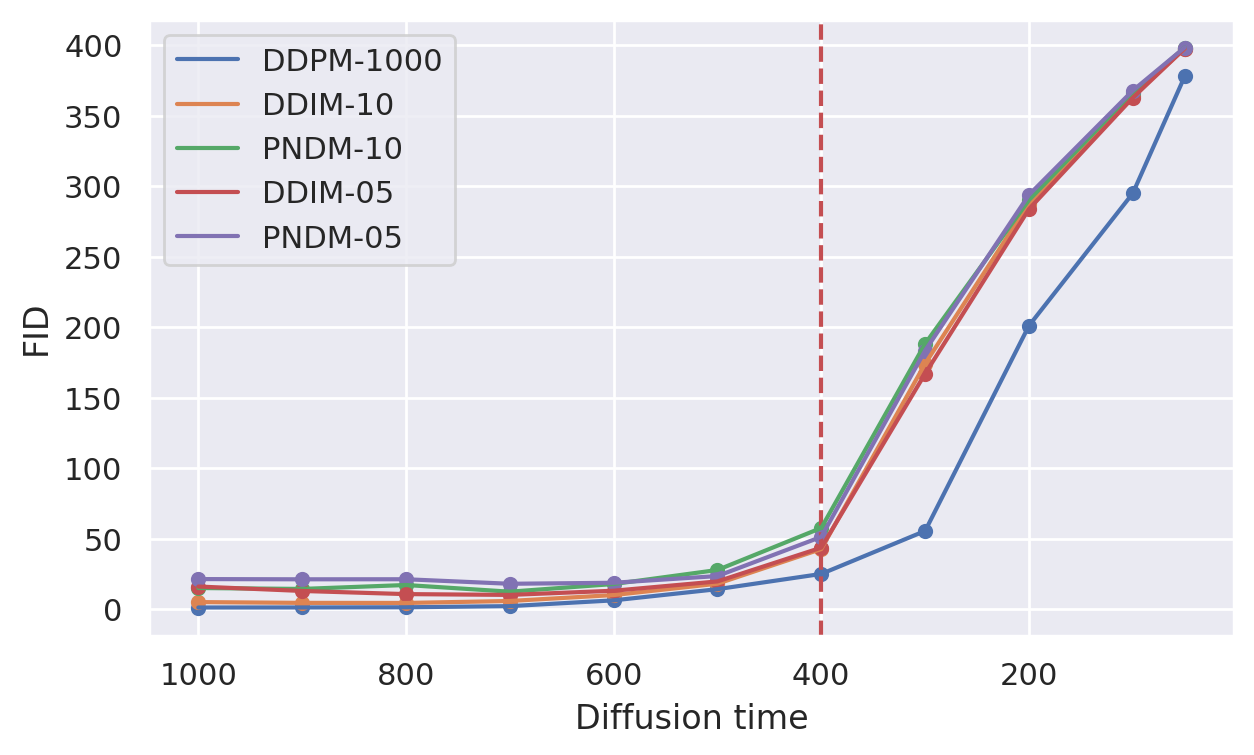}
    \subcaption{MNIST}
  \end{minipage}\vfill
  \begin{minipage}[b]{\textwidth}
    \centering
    \includegraphics[width=\textwidth]{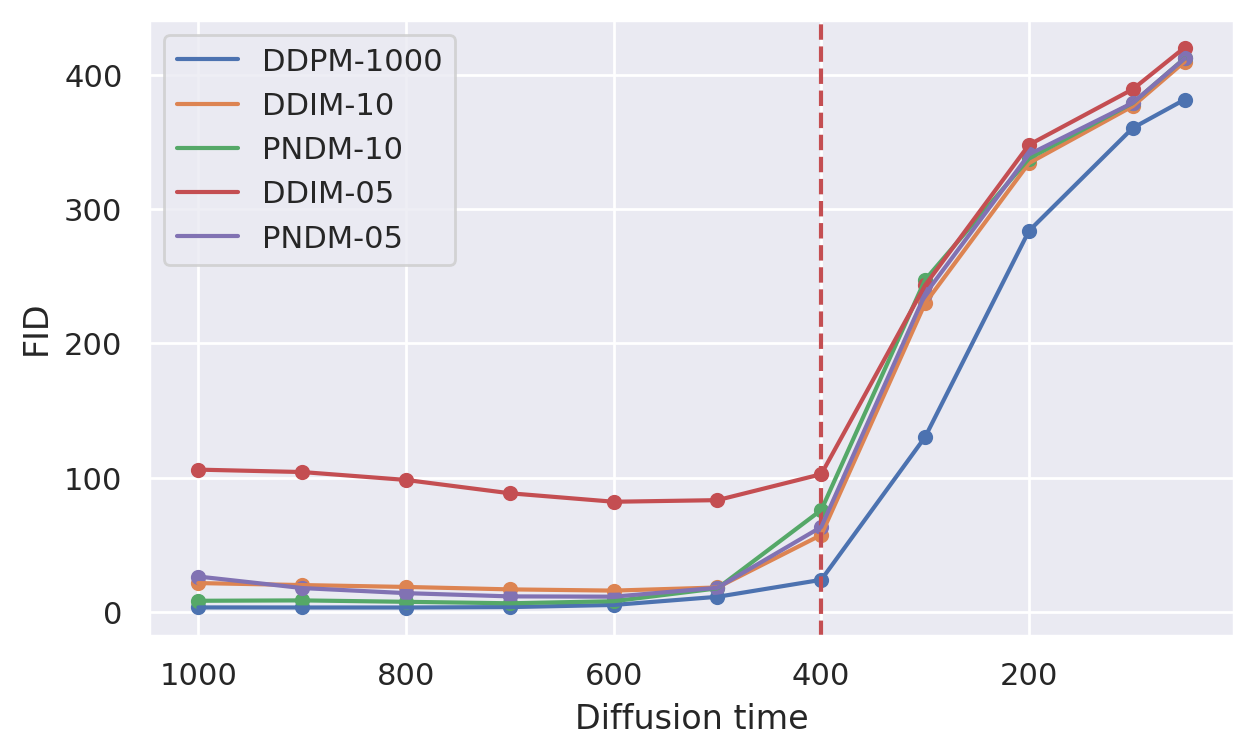}
    \subcaption{CIFAR10}
  \end{minipage} 
\end{minipage}%
\begin{minipage}[b]{.34\textwidth}
  \begin{minipage}[b]{\textwidth}
    \centering
    \includegraphics[width=\textwidth]{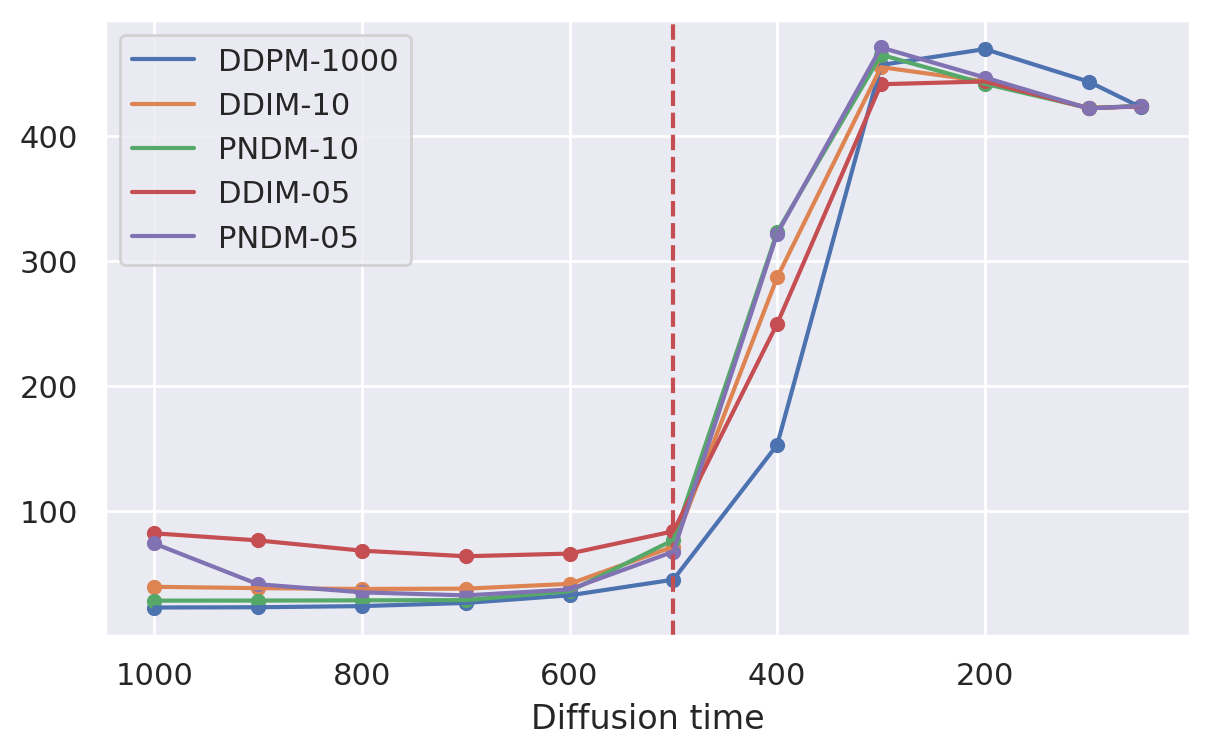}
    \subcaption{Imagenet64}
  \end{minipage} \vfill
  \begin{minipage}[b]{\textwidth}
    \centering
    \includegraphics[width=\textwidth]{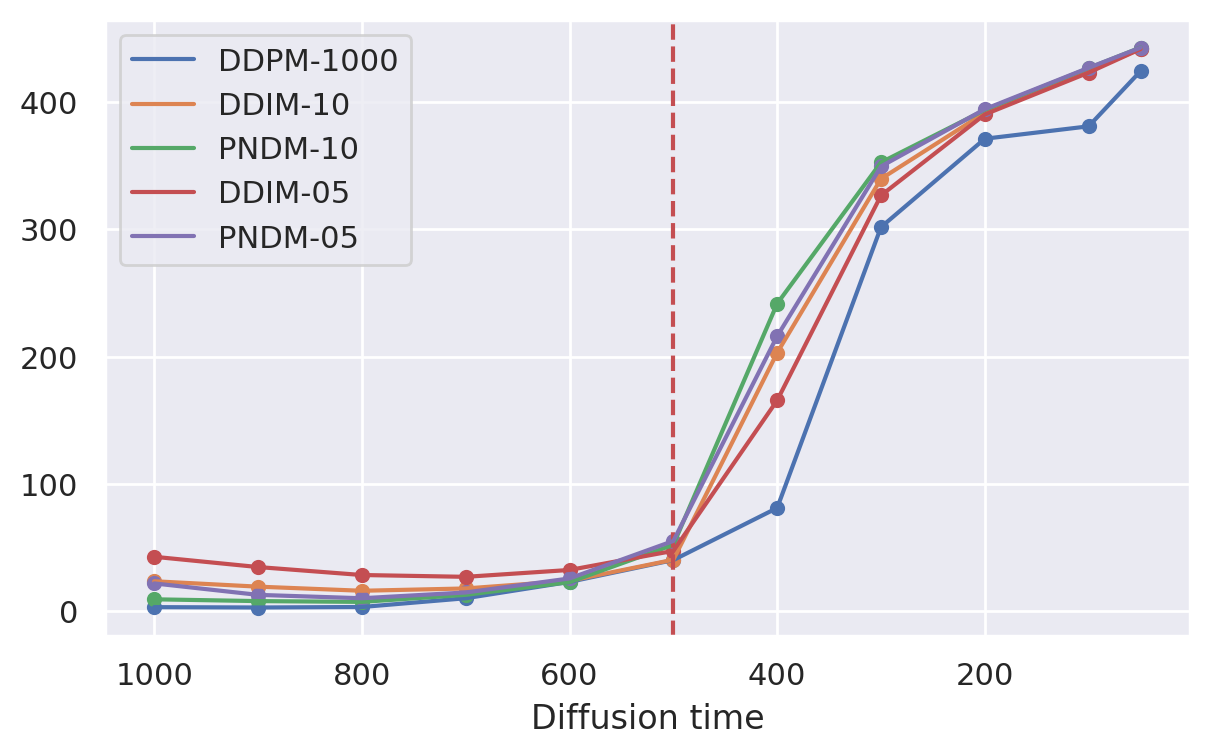}
    \subcaption{CelebA64}
  \end{minipage}
\end{minipage}%
\begin{minipage}[b]{.298\textwidth} 
  \includegraphics[width=\textwidth]{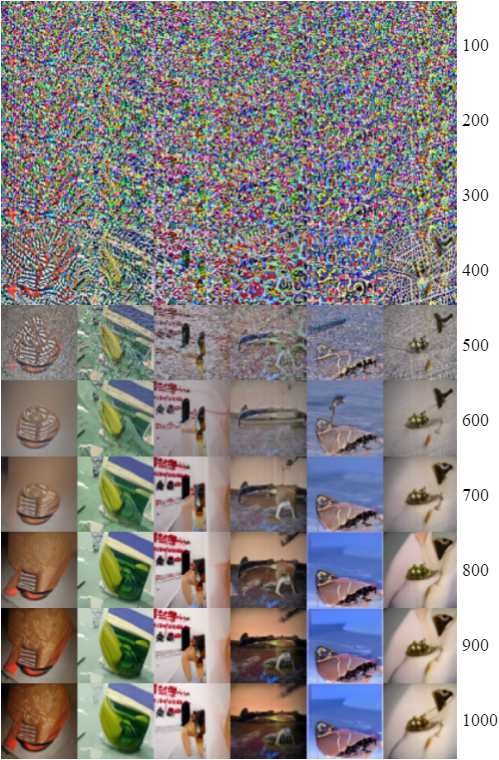}\vspace{.3\baselineskip}
  \subcaption{Imagenet late start generation} \label{fig: FID results}
\end{minipage}%
\caption{Figure taken from \citep{raya2023spontaneous}. Analysis of the model's performance, as measured by FID scores, for different starting times using three different sampling methods: the normal DDPM sampler with decreasing time steps from $T=1000$ to 0, and fast sampler DDIM and PSDM for 10 and 5 denoising steps. The vertical line corresponds to the maximum of the second derivative of the FID curve, which offers a rough estimate of the first bifurcation time. (e) Illustrates samples generation on Imagenet64, while progressively varying the starting time from 1000 to 100.}
\label{fig:late_start_analysis}
\end{figure} 

\section{Experimental evidence of phase transitions in trained diffusion models}
The presence of one or more phase transitions in generative diffusion models is hard to prove theoretically for complex data distributions. However, symmetry breaking can be inferred experimentally from trained networks. For example, \cite{raya2023spontaneous} showed that the generative performance of models trained on images stay largely invariant when the reversed dynamics is initialized before $t_\text{end}$ up to a critical point (see Fig.\ref{fig: FID results}), as far as the system is initialized with a Gaussian with properly chosen mean vector and covariance matrix. This is consistent with out theoretical analysis since, prior to the first phase transition, the marginal distributions have a single global mode and are well approximated by normal distributions. The effect of this form of symmetry breaking was further studied both experimentally and theoretically in \cite{sclocchi2024phase}, where it was found that the generative diffusion models trained on natural images undergo a series of phase transitions corresponding to hierarchical class separations, where low-level visual features emerge at earlier times of the diffusion process while higher level semantic features emerge later. These results where further studied in \cite{biroli2024dynamical}, where the authors provided a series of analytic formulas for the critical times and verified the prediction experimentally on several image datasets. In particular, it was found that symmetry breaking time corresponding to splits in different classes (e.g. images of dogs and cats) can be predicted based on the eigenvectors of the covariance matrix of the data. Together, these results strongly suggest that generative diffusion models undergo symmetry breaking phase transitions under most realistic data distributions. 

\section{Conclusions}
In this paper, we presented a formulation of generative diffusion models in terms of equilibrium statistical mechanics. This allowed us to study the critical behavior of these generative models during second-order phase transitions and the disordered thermodynamics of models trained on finite datasets. Our analysis establishes a deep connection between generative modeling and statistical physics, which may in the future allow physicists to study these machine learning models using the tools of computational and theoretical physics.

\bibliographystyle{apalike}
\bibliography{bibliography.bib}






\end{document}